\documentclass[11pt,a4paper]{article}
\usepackage[hyperref]{naaclhlt2018}
\usepackage{times}
\usepackage{multirow}
\usepackage{latexsym}
\usepackage{graphicx}
\usepackage{wrapfig}

\usepackage{url}

\aclfinalcopy 

\title{Towards Verifying Semantic Roles Co-occurrence}

\author{Aliaksandr Huminski \\
  Institute of High Performance \\
  Computing, Singapore\\
  {\tt huminskia@}\\
  {\tt ihpc.a-star.edu.sg} \\\And
  Hao Zhang\\
  AI Initiative \\
  A*STAR, Singapore\\
  {\tt zhang\underline{ }hao@}\\
  {\tt scei.a-star.edu.sg} \\\And
  Gangeshwar Krishnamurthy \\
  AI Initiative\\
  A*STAR, Singapore \\
  {\tt gangeshwark@}\\
  {\tt ihpc.a-star.edu.sg} \\}

\date{}

\begin{document}
\maketitle
\begin{abstract}
  Semantic role theory considers roles as a small universal set of unanalyzed entities. It means that formally there are no restrictions on role combinations. We argue that the semantic roles co-occur in verb representations. It means that there are hidden restrictions on role combinations. To demonstrate that a practical and evidence-based approach has been built on in-depth 
  analysis of the largest verb database VerbNet. The consequences of this approach are considered.
\end{abstract}

\section{Introduction and Motivation}
50 years ago sharp the famous article "The case for case" was published \cite{fillmore1968case}. It gave a start to the semantic role theory that is widely used in QA systems, machine translation, summarization and information extraction.
Meanwhile, even now, after so long period of time and hundreds of publications, yet there is no consensus regarding 2 points that are considered as main drawbacks of the theory:
\begin{enumerate}
\item What is the list of roles (number of roles and their definitions)? And a related question: How fine semantic roles should be?
\item Is there any internal organization inside of the list of roles? In other words, is the list of roles structured?
\end{enumerate}
We investigate the point \#2 through in-depth analysis of the largest verb database VerbNet.\\
The content of the paper is structured as follows. In section \ref{sec:related-work} we describe the related work. In section \ref{sec:relationtypes} we define two types of relations that are supposed to be verified across roles. Then, in section \ref{sec:dataset} the database is described for getting role representation as a vector. In section \ref{sec:analysis} three methods are applied to validate if there are relations across the role vectors. Finally, in section \ref{sec:conclusion}, we present conclusion, consequences and the plans for future research in this area.

\section{Related Work}
\label{sec:related-work}
The question about internal organization inside of the list of roles can be rephrased in a more general form: are there any types of relations across roles? 
Many linguists pointed out that roles are related in some way. They paid attention to the fact that some semantic roles co-occur while the others do not. For example, the role frame \{Agent, Patient, Instrument\} represents lots of verbs but the role frames \{Patient, Experiencer\} or \{Beneficiary, Goal\} do not represent any verb at all. In FrameNet \cite{fillmore2002framenet} there are 2 types of role-to-role relations -- "requires" and "excludes" -- that indicate for a specific role in a frame which roles are required and which are excluded. 

Using semantic role theory, it is hard to give any explanation for that. Nevertheless, a list of roles is considered as unstructured and this leads to two consequences.

First, there is not any procedure or method to distinguish a possible frame of roles that can be associated with a single verb from an impossible one. In other words, there is no way of allowing the natural groups of semantic roles and disallowing the unnatural ones \cite{levin2005argument}. 

The second consequence follows from the first one: if it is unable to explain why some frames of roles are valid while the others are not, the obvious conclusion is that roles are not primitive elements in semantic representation. They are not unanalyzable \cite{rappaport1988theta}. Treating semantic roles as primitives that are unrelated to one another makes valid role combination purely accidental \cite{davis2000linking}.

\section{Types of Relations between Roles to Be Verified}
\label{sec:relationtypes}

In our investigation we will consider if the following two types of relations take place across role frames.
\begin{itemize}
\item \textbf{\em{Occurrence}:} a role A occurs only if a role B (or a set of roles B,C,..) occurs. Occurrence is a one-way relation: if an appearance of a role A depends on an appearance of a role B, it doesn't mean a role B depends on a role A.
\item \textbf{\em{Co-occurrence}:} a role A and a role B occur in the context of each other. Co-occurrence of roles means their mutual dependency.
\end{itemize}

\section{Extraction and Vectorization of Roles from VerbNet}
\label{sec:dataset}
We will verify the existence of two types of relations on VerbNet (VN). VN was chosen because of the following 4 reasons.

First, VN is the largest domain-independent computational verb lexicon currently available for English \cite{schuler2005verbnet}. It is extensive enough to be used as a platform for running our algorithms. 

Second, it provides semantic role representation for all verbs from the lexicon. 

Third, the roles used in VN are not so fine-grained and frame-specific as in FrameNet and not so coarse-grained and compact as in Propbank \cite{palmer2005proposition}. 

Fourth, VN role set was considered as a base for creation of ISO Standard -- SemAF-SR -- for semantic role annotation \cite{petukhova2008lirics,bunt2013conceptual}.

In this paper, we use the latest VerbNet version 3.2b that contains 6394 verbs.

\subsection{Discrepancy between Real and Possible Number of Role Frames}
A role frame for a verb in VerbNet contains min. 1 role and max. 6 roles. Having $n = 30$ roles total in VN, the number of all possible role frame combinations (FC) is calculated as the number of subsets of the size $1 \leq k \leq 6$:

\begin{equation}
FC = \sum_{k=1}^6 {30\choose k} = \sum_{k=1}^6 \frac{30!}{(30-k)! k!} = 768211
\end{equation}

In reality there are only 107 unique role frames in VN. This fact demonstrates a large discrepancy between the number of "natural combinations" and the number of all possible combinations of role frames, and indicates there are some hidden restrictions on role combinations.

\subsection{Extraction of Verb Classes and Corresponding Role Frames}

To analyze the relations across roles, we need first to extract the verbs with corresponding role frames from VN. The verbs are organized through 498 verb classes: 277 root verb classes (RVC) and 221 sub-root verb classes (sub-RVC). Each RVC and sub-RVC has its own role frame. For example, RVC \emph{destroy-44} has the role frame \{Agent, Patient, Instrument\}. It is a common situation when sub-RVC inherits roles from RVC and has the same role frame. In this case, sub-RVC can be merged with their parent RVC. We found only 13 sub-RVCs that have their own additional roles besides the roles inherited from parent RVCs.

To access VN we use JVerbNet toolkit 1.2.0. \cite{jverbnet2012}. To extract all RVCs and sub-RVCs that are different from their parent RVCs, we use Depth-First-Search method \cite{tarjan1972dfs} that allows making recursive search from RVC to its sub-RVCs.

As a result, we compressed 498 verb classes and reduce them into 290 verb classes (277 RVCs and 13 sub-RVCs). For these 290 classes we extracted the name of the class, the number of its members and the roles in the class frame. Example: class \emph{break-45.1}; members: 24; role frame: \{Agent, Patient, Instrument, Result\}.

\subsection{Roles as Vectors}
\begin{table*}[t!]
\centering
\begin{center}
\begin{tabular}{|l|c|c|c|c|c|c|c|c|c|}
\hline \bf Classes & \bf Roles & \bf Members & \bf Agent & \bf .. & \bf Patient & \bf .. & \bf Instr. & \bf .. \\ \hline

destroy-44 & Agent+Patient+Instr. & 31 & 1 & .. & 1 & .. & 1 & .. \\ \hline
pelt-17.2 & Agent+Theme+Destination & 7 & 1 & .. & 0 & .. & 0 & .. \\ \hline
cost-54.2 & Theme+Value+Beneficiary & 5 & 0 & .. & 0 & .. & 0 & .. \\ \hline
break-45.1 & Agent+Patient+Instr.+ Result & 24 & 1 & .. & 1 & .. & 1 & .. \\ \hline
.. & .. & .. & .. & .. & .. & .. & .. & .. \\ \hline
\end{tabular}
\end{center}
\caption{\label{group-table} Representations of the verb classes and the roles. (Instr. = Instrument)}
\end{table*}

Since there are 30 different roles in VerbNet, we constructed a 30-dimensional vector for each extracted verb class. Each element of the vector represents one role: if a verb class has a role in its frame, the corresponding position is set as 1, otherwise 0. The result is shown in the Table \ref{group-table}.

If each verb class is represented by 30-dimensional vector, each role is represented by 290-dimensional vector according the number of classes.

The last step is to represent a role not as a vector based on classes but as a vector based on verbs. A class contains a group of verbs that share the same role frame. So, to expand a vector on verbs we need just to expand 1s or 0s $n$ times, where $n$ is the number of members of a class. Using this way, each role is transformed from 290-dimensional vector into 6394-dimensional vector.

\section{Analysis of Role Vectors}
\label{sec:analysis}

\begin{table*}[t!]
\centering
\begin{center}
\begin{tabular}{|l|c|c|c|c|c|c|c|c|c|c|}
\hline \bf Roles & \bf Agent & \bf ... & \bf Beneficiary & \bf Co-Agent & \bf ... & \bf Exp. & \bf Instr. & \bf ...  & \bf Material& \bf ... \\ \hline

Agent & 100 & ...& 6.2 & 3.1 &... & 0.5 & 20.6& ...  & 2.4 &...\\ \hline
... &... & ...&... &... &... &...  &... &... &... &... \\ \hline
Beneficiary & 95.7 &... & 6.2 & 3.1 &... & 20.7 & 16.2 & ... & 24 &...\\ \hline
Co-Agent & 100 &... & 8.7 & 100 &... & 0 & 0 & ...& 0  & ...\\ \hline
... &... &... &... &... &... &...  &... &... &... &...\\ \hline
Exp. & 4.6 &... & 0 & 0 &... & 100 & 4.6 &... & 0 &... \\ \hline
Instr. & 100 &... & 0 & 0 &... & 2.7 & 100 &...& 0  &... \\ \hline
... &... &... &... &... &... &...  &... &... &... &...\\ \hline
Material & 100 &... & 65 & 0 &... & 0 & 0 &...& 100  &... \\ \hline
... &... &...&... &... &... &...  &... &... & ...&...\\ \hline
\end{tabular}
\end{center}
\caption{\label{dep-table} Results of the roles occurrence. (Exp. = Experiencer; Instr. = Instrument) }
\end{table*}
 
\begin{table*}[t!]
\centering
\begin{center}
\begin{tabular}{|l|c|c|c|c|c|c|c|c|c|}
\hline
& \bf k=29 & \bf k=28 & \bf k=27 & \bf ... & \bf k=2 \\ \hline
\multirow{6}{2.9em}{Clusters} & Cluster "Experiencer  & Cluster "Experiencer & Cluster "Experiencer & .. & Cluster "Agent\\ 
& \& Stimulus" & \& Stimulus" & \& Stimulus" &  & \& Theme"\\
& + Clusters with one & + Cluster "Material  & + Cluster "Material & .. & + Clusters with \\ 
& role in each & \& Product" & \& Product" & & other roles\\ 
& & + Clusters with one & + Cluster "Agent & ..& clustered\\
& & role in each & \& Theme" &&\\
& & & + Clusters with one &&\\
& & & role in each && \\
\hline
\end{tabular}
\end{center}
\caption{\label{tab:clusters-table} Results of the role vectors clustering. }
\end{table*}

Three methods have been applied to the analysis of 6394-dimensional role vectors.

\subsection{Analysis of Occurrence}

The method for calculation of contextual occurrence includes two steps:

\begin{enumerate}
\item For a pair of 2 roles, such as "Agent-Instrument", we get their vectors and count the number of positions with "1" for both vectors: $val\textsubscript{$common$}$.
\item After computing $val\textsubscript{$common$}$, we count all "1" for the first vector (Agent): $sum\textsubscript{$agent$}$, and all "1" for the second vector (Theme): $sum\textsubscript{$theme$}$. Then we compute the ratio $P(Theme|Agent) = \frac{val _{common}}{sum _{agent}}$ and the ratio: $P(Agent|Theme) = \frac{val _{common}}{sum _{theme}}$.
\end{enumerate}

We apply this method to all the role pairs. The results are shown in the Table \ref{dep-table}.

One can see from the Table \ref{dep-table} that Co-Agent, Instrument and Material occur $only$ in the context of Agent (100\%), Beneficiary occurs 95.8\% in the context of Agent. Agent is almost independent from all of them: 6.2\% occurrence in the context of Beneficiary, 3.1\% with Co-Agent, 20.7\% with Instrument and 2.4\% with Material.

\subsection{Analysis of Co-occurrence}

For analysis of co-occurrence we use 2 methods. Before being applied, role vectors are transformed the following way: for each vector a position with "1" remains the same while a position with "0" is replaced with a random uniform numbers in the range (-1.0, 1.0) to mimic the Word2Vec \cite{word2vec} model for representing a verb.

\subsubsection{Visualization of Roles Co-Occurrence via t-SNE Dimensionality Reduction}

t-Distributed Stochastic Neighbor Embedding (t-SNE) \cite{maaten2008visualizing} is a machine learning technique for dimensionality reduction that is particularly well suited for the visualization of high-dimensional datasets.

\begin{figure}
\centering
\includegraphics[scale=0.28]{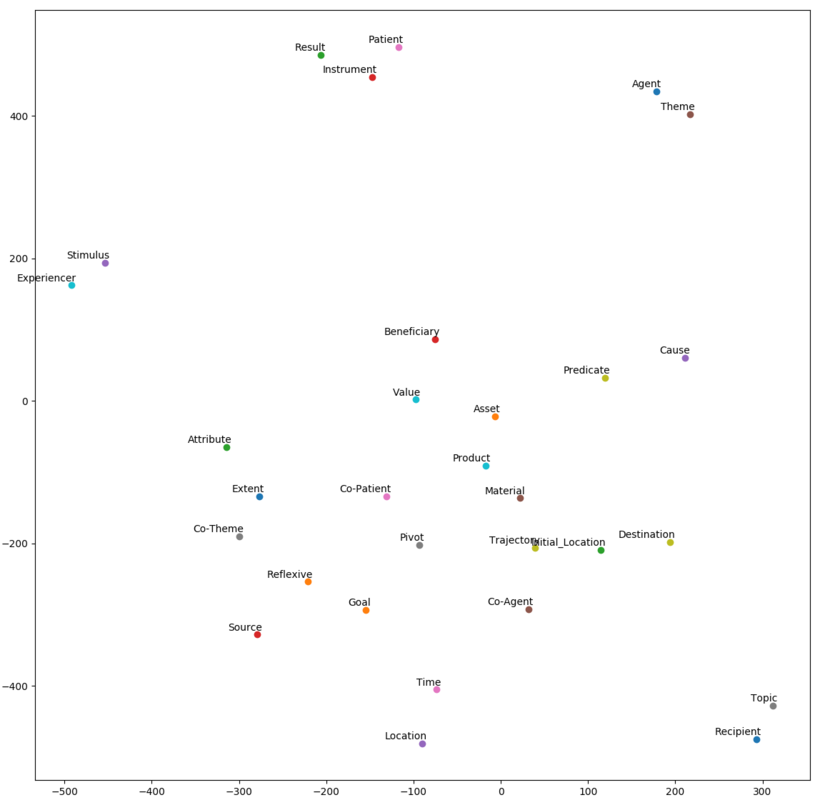}
  \caption{Visualization of Co-occurrence of the Roles.}
  \label{fig:tsne_result}
\end{figure}

In this paper, we use t-SNE for reducing role representations into 2-dimensional space, which can be easily visualized in a scatter plot, as shown in the Figure \ref{fig:tsne_result}. Some roles are very close to each other: "Agent-Theme", "Experiencer-Stimulus", "Topic-Recipient". Proximity of the roles represents here not their similarity but their co-occurrence in the role frames.

\subsubsection{Visualization of Roles Co-Occurrence via Clustering}

K-means clustering \cite{kanungo2002kmeans} is an unsupervised technique to partition $n$ data points into $k$ clusters where $k < n$. We use this technique in order to reveal the co-occurrence of certain roles.

Having 6394 verbs and 30 roles, the original data matrix of size $30 \times 6394$ is created, where each role has a vector representation of size 6394. After that, the dimension of the vector is reduced to 30 using Principal Component Analysis transformation \cite{pca1901}. K means clustering is performed with $k = 29$ to find the cluster with 2 roles.

On the next step, we decreased the cluster size $k$, by 1 to find the clustered roles with $k = 28$ and so on until $k = 2$. The results for $k = 29, 28, 27,$ and $2$ are presented in the Table 3.

One can see from the table that the strongest tie among 30 roles exists between Experiencer and Stimulus ($k = 29$). On the 2nd and 3rd places regarding the tie strength ($k = 28$ and $k = 27$) are relation "Material-Product" and "Agent-Theme". Also the clustering when $k = 2$ shows that Agent and Theme together are far away from the other roles.

\section{Conclusion, Consequences and Future Work}
\label{sec:conclusion}

In this paper we have described 3 methods: calculation of contextual occurrence, t-SNE Dimensionality Reduction and K means clustering. Being applied to VerbNet role vectors, they demonstrate that the roles are related.

Two important consequences follow from this demonstration. If a role occurs in the context of another one or if there is a co-occurrence of 2 roles, then:
\begin{itemize}
\item A role definition is supposed to be re-considered: it is not just a relation between an argument and a verb.
\item A minimal role frame for a verb is supposed to consist of two roles, not one role.
\end{itemize}
We are going to investigate these two points as two hypotheses for verification in future.

\bibliography{naaclhlt2018}
\bibliographystyle{acl_natbib}

\end{document}